\documentclass{article}

\usepackage{color, soul}
\usepackage[numbers]{natbib}  
\usepackage{amsfonts} 
\usepackage{amsmath}
\usepackage{graphicx}
\usepackage{csvsimple}
\usepackage{float}
\usepackage{subfig}
\usepackage[margin=1in]{geometry}
\usepackage{dirtytalk}

\title{A General, Evolution-Inspired Reward Function for Social Robotics}
\author{Thomas Kingsford\\University of Auckland\\tkin063@aucklanduni.ac.nz}

\begin{document}

\maketitle

\begin{abstract}
The field of social robotics will likely need to depart from a paradigm of designed behaviours and imitation learning and adopt modern reinforcement learning (RL) methods to enable robots to interact fluidly and efficaciously with humans. In this paper, we present the \textit{Social Reward Function} as a mechanism to provide (1) a real-time, dense reward function necessary for the deployment of RL agents in social robotics, and (2) a standardised objective metric for comparing the efficacy of different social robots. The \textit{Social Reward Function} is designed to closely mimic those genetically endowed social perception capabilities of humans in an effort to provide a simple, stable and culture-agnostic reward function. Presently, datasets used in social robotics are either small or significantly out-of-domain with respect to social robotics. The use of the \textit{Social Reward Function} will allow larger in-domain datasets to be collected close to the behaviour policy of social robots, which will allow both further improvements to reward functions and to the behaviour policies of social robots. We believe this will be the key enabler to developing efficacious social robots in the future.
\end{abstract}


\section{Introduction}

\paragraph{} Social robots aim to achieve tasks by interaction and collaboration with humans on a social level. This requires an understanding of social skills. But what are social skills? Human social skills are a means-to-an-end to facilitate communal and societal cooperation and have enabled humans to become the dominant species on Earth.

\paragraph{} The field of social robotics has yet to leverage the fast-paced advancement that both broader robotics and machine learning have benefited from in recent years. We propose there are two missing pieces:

\begin{enumerate}
\item Reinforcement learning

\item Objective evaluation mechanisms
\end{enumerate}

\paragraph{} The first missing piece is the realisation that social robotics is a sequential decision making problem, and should be addressed using modern reinforcement learning approaches. This allows us to escape the limitations of systems designed by hand or by mimicry.

\paragraph{} How do humans learn social skills? Humans can be thought of as a meta-learning system. Evolution (as a learning algorithm) has endowed individual humans with the ability to learn based on experience. This too is true of humans' social skills. There is clearly poverty of stimulus. Humans aren't merely general learners which are dropped into society as infants and learn everything from scratch. There is clearly a prior or a bias that is used to bootstrap learning. Humans have intrinsic desires, which encourage learning early in life. Some of these desires change as the human infant develops, and others remain. It is these intrinsic desires which enable human learning, including the learning of social skills, and which may be thought of as a reward function.

\paragraph{} The difficulty here is of course that it is difficult for robots to directly learn social behaviours. A significant contributor to this difficulty is that simulation is largely infeasible in a social robotics context. While social robots themselves can be simulated, their environment (the humans they interact with) cannot be. We propose that, if it is possible for robots to learn social skills, it will be possible only by providing them with the same genetically endowed social cues as humans. Any less, the problem is too sample inefficient to be tractable. Any more and the problem is likely to be over-prescriptive of a particular solution, with no guarantees of the degree of optimality of that particular solution. Moreover, it will be important to provide them with the opportunity to learn in long-term real-world social deployments to accrue sufficient data.

\paragraph{} If we wish for autonomous systems, such as social robots, to interact collaboratively and fluidly with humans in a social context, we have three options:

\begin{enumerate}
\item Intricately design their interactions, based on engineers' understanding

\item Learn social behaviour as a byproduct of optimising with respect to collaborative tasks. This is akin to robotic learning taking the learning responsibilities of the individual human \textit{and} evolution at large.

\item Bootstrap by providing fundamental social capabilities as a reward signal in training, thus mimicking the learning environment genetically endowed upon humans, then learn. This is akin to robotic learning being restricted to the learning responsibilities of the individual human \textit{only}, but not evolution.
\end{enumerate}

\paragraph{} We propose that rapid development in this field is feasible only by adopting this latter approach. By treating social robotics as a sequential decision making problem, solving it with modern reinforcement learning algorithms, but starting with capabilities as close to those genetically endowed upon humans as possible, we believe social robotics can begin to experience the rapid advancement seen in other fields of machine learning and robotics.

\paragraph{} The second missing piece is the development of objective evaluation mechanisms. This has been exceedingly hard in social robotics. This would enable both standardised evaluations of systems (there is no current agreement of how to objectively and quantitatively evaluate a social robot's performance) and reinforcement learning. Such objective evaluation mechanisms must be automatic and low latency, 

\paragraph{} We propose the \textit{Social Reward Function}, a real-time real-valued reward function, designed to mimic to the best of our abilities those fundamental social capabilities endowed upon humans by evolution, and which can be used for both learning and evaluation of social robotic systems. An implementation is made publicly available as a Python library\footnotemark.

\footnotetext{https://github.com/TomKingsfordUoA/social-reward-function}

\section{Designing a reward function}

\paragraph{} The design problem is to produce a sufficiently dense reward function based on our best guess of genetically endowed social cues. In principle, we want to avoid over-engineering the reward function as this amounts to \say{assuming a solution} and might result in optimal behaviour with respect to the designed reward function failing to be sufficiently socially efficacious. However, we simultaneously need to ensure there is sufficient information content in the reward signal that learning is tractable.

\paragraph{} So, how do we determine the appropriate genetically endowed capabilities of humans? Two general approaches can be taken. 

\paragraph{} In the first, the development of humans in pre-adulthood is studied. Since infants have been exposed to minimal experience, it is likely that the capabilities they demonstrate are largely evolutionarily-endowed. Human infants are born in a significantly immature state and don't reach physical and intellectual maturation for more than two decades. This is in stark contrast to many other animal species. This presents a risk in that social skills which are truly genetically endowed may be indistinguishable from those which are learned as biological maturation occurs in concert with experiential learning. Nonetheless, if we focus on studying the social skills of humans in infancy rather than childhood and beyond, the effect of experiential learning can be minimised and observed effects can be assumed to be attributable directly to evolution.

\paragraph{} In the second, humans are studied across cultures for common social skills. It is likely that culture-agnostic social skills are either endowed by evolution or are learned but so general that their incorporation in a reward function doesn't present a risk of over-engineering.

\paragraph{} \cite{Kingsford2021} conclude the following social capabilities are likely to be genetically endowed in humans:

\begin{enumerate}
\item \say{Face and face-like objects detection \citep{Mumme2001, Sirois2007}, including facial expression recognition \citep{Farroni2007}. The \textit{generation} of certain facial expressions is likely innate \citep{Sullivan2003} and could suggest the detection of those facial expressions, but not others, is innate.}

\item \say{Eye-like object detection and limited gaze following \citep{SotoIcaza2015}.}

\item \say{Proprioceptive mimicry \citep{Mumme2001, SotoIcaza2015}.}

\item \say{Biological motion detection \citep{SotoIcaza2015}.}

\item \say{Prenatal maternal/foetal physiological bond yielding vocal emotion recognition in the neonate \citep{Mastropieri1999}.}

\end{enumerate}

\paragraph{} For a more thorough review of social cognition in the Psychology literature and its relevance to social robotics, the reader is referred to \cite{Kingsford2021}.

\section{Components}

\paragraph{} For the purposes of designing the \textit{Social Reward Function}, the following capabilities are considered evolutionarily-endowed in humans, amenable to implementation in a robot, and able to be described by a reward function:

\begin{enumerate}
\item Simple facial expressions (more advanced facial expressions are likely learned and may be culture-dependent)

\item Emotion in voice (although this is likely learned in the womb by physiological connection with mother)

\item Touch is favourable \textit{ceteris paribus}

\item Interaction is favourable \textit{ceteris paribus}
\end{enumerate}

\paragraph{} We have designed the \textit{Social Reward Function} to incorporate simple facial expressions and interaction principally, with voice emotion secondarily. Modern machine learning models for Facial Emotion Recognition (FER) are robust to real-world situations, while models for Speech Emotion Recognition (SER) often struggle to generalise across domains. It is likely that this is due to the dominant datasets for SER suffering from cultural homogeneity and from being collected from actors in laboratory settings. It is hoped that the collection and annotation of more realistic datasets (perhaps facilitated by widespread deployments of social robots) will lead to the development of more robust SER models.

\paragraph{} It is likely that touch can be incorporated in future works. Touch has been used by prior art as a component of the reward signal in RL for social robots \citep{Qureshi2017}. However, it is presently unclear how general this is (both in terms of situations and robot morphology) and more data must be collected. For instance, particular scenarios or robot morphologies might unfairly encourage touch from which inappropriately inflated social value scores might be inferred. It is likely that a specification for touch sensors and robot morphology will need to accompany the inclusion of touch in this reward function.

\paragraph{} A survey of FER and SER models for which publicly-available implementations were available was conducted. Those that had reasonable performance on datasets and passed subjective assessments of robustness were included in the \textit{Social Reward Function}. A summary of the included models is presented in Table \ref{table:component_models}. 

\begin{table}[H]
\centering
\begin{tabular}{|p{6cm} p{1.5cm} p{3cm} p{3cm}|} 
 \hline
 Model Name & Modality & Datasets & Test-Set Accuracy\footnotemark \\ [0.5ex] 
 \hline\hline
 Residual Masking Network \citep{residual_masking_network_2020} & FER & FER2013 \citep{goodfellow2013} & 69\% (76.82\%) \\ 
 MevonAI \citep{mevonai_2020} & SER & RAVDESS \citep{Livingstone2018} & 34\% (66.0\%) \\
 Emotion Recognition Using Speech \citep{emotion_recognition_using_speech_2020} & SER & RAVDESS \citep{Livingstone2018} \par TESS \citep{tess_2010} \par EMODB \citep{Burkhardt2005} &  34\% (79.5\%) \\ 
 Multi-Modal SER \citep{sahu2019multimodal} & SER & IEMOCAP \citep{iemocap} & 33\% (56.6\%) \\[1ex]  
 \hline
\end{tabular}
\caption{\textit{Social Reward Function} Constituent Models}
\label{table:component_models}
\end{table}

\footnotetext{Accuracy according to our experimental results, based on FER2013/RAVDESS partitioned into training and test sets such that no actor is present in both datasets. Author-published accuracy (based on whatever dataset was used by the authors) is presented in parentheses.}

\paragraph{} The above models are combined to produce two emotion perception matrices: $\bar{X}_\text{FER} \in \mathbb{R}^{n_\text{FER} \times k}$ and $\bar{X}_\text{SER} \in \mathbb{R}^{n_\text{SER} \times k}$, where there are $n$ models estimating $k$ emotions.

\begin{eqnarray}
\bar{X} &:=& [\vec{x}_1 | \vec{x}_2 | ... | \vec{x}_n]^T\\
\vec{x}_i^T \vec{x}_i &=& 1
\end{eqnarray}

\paragraph{} For each modality, per-model vectors are averaged to produce per-modality vectors.

\begin{eqnarray}
\vec{x}_\text{FER} &:=& \frac{1}{n_\text{FER}} \sum{\vec{x}_i^\text{FER}}\\
\vec{x}_\text{SER} &:=& \frac{1}{n_\text{SER}} \sum{\vec{x}_i^\text{SER}}
\end{eqnarray}

\paragraph{} The reward function is then calculated at each time step as follows, where $k_\text{FER}, k_\text{SER}, \vec{w}_\text{FER}, \vec{w}_\text{SER}$ are design parameters:

\begin{eqnarray}
r &:=& k_\text{FER} \vec{w}_\text{FER}^T \vec{x}_\text{FER} + k_\text{SER} \vec{w}_\text{SER}^T \vec{x}_\text{SER} \in \mathbb{R}\\
\vec{w}_\text{FER}^T \vec{w}_\text{FER} = \vec{w}_\text{SER}^T \vec{w}_\text{SER} &=& 1\\
k_\text{FER}, k_\text{SER} &\in& \mathbb{R}\\
\vec{w}_\text{FER}, \vec{w}_\text{SER} &\in& \mathbb{R}^k
\end{eqnarray}

\section{Evaluation method}

\paragraph{} Two approaches are appropriate for evaluating the correctness of the \textit{Social Reward Function}: direct and indirect evaluation.

\paragraph{} A direct indicator of the correctness of the \textit{Social Reward Function} is an evaluation of the correlation between the function's assessment of social situations, and that of a human observer acting as an oracle. This approach allows large amounts of data to be collected by annotating previously-collected scenes across a variety of scenarios. The downside is that scenarios are not constrained to the human-robot-interaction domain and hence might not accurately characterise correctness in that domain.

\paragraph{} An indirect indicator is to expose human participants to a selection of robot agents and observe their interactions. It is important that such an experiment be focused on human-robot interaction, but that the participants be \textit{blinded} (i.e. the human participants shouldn't be aware their responses to the robot are being observed). If the experiment is not blinded, it is likely the participants will modify their own reactions either consciously or subconsciously. If the experiment is not focused on human-robot interaction, the data can be considered out-of-domain and might not generalise to the prediction of human emotions in a human-robot interaction domain. In such an experiment, (non-participant) human observers and the \textit{Social Reward Function} evaluate the emotional reaction of the human participants during the experiment, and the consistency of these evaluations is assessed. Such an evaluation is highly domain-specific to human-robot-interaction, but will contain relatively limited quantities of data.

\paragraph{} In both cases it is notable that humans are often fallible witnesses and often not consciously aware of their true emotional responses to social situations, hence their responses might contain inaccuracies. Moreover, these inaccuracies might be systematic and displayed across different instances of similar types of scenarios and potentially across different human observers.

\section{Results}

\subsection{Performance of components in isolation}

\paragraph{} Before evaluating the \textit{Social Reward Function} as a whole, it is worthwhile to characterise the performance of the component models on relevant FER and SER datasets. To allow for a direct comparison of model architectures without the effects of different datasets, we re-train each model on FER2013 or RAVDESS only (for FER and SER models respectively). Specifically, datasets were split into training and test sets by partitioning on actors, and not on individual samples. In this way, the test set provides a more challenging but more realistic test of the ability of the models to generalise to unseen persons, as would be required in a robotics application.

\paragraph{} Since it is possible that pre-trained models were exposed to test set data which would invalidate test set results, models were trained from random initial parameters on the training set only, then evaluated on both the training and test sets. Detailed results are presented in Appendix \ref{appendix_component_model_results}. 

\paragraph{} The FER model performs well, producing a top-1 accuracy of $69\%$. Unfortunately, the SER models all produce a top-1 accuracy of approximately $33\%$. This is likely due to the small dataset size (both in terms of number of samples and diversity of actors and scenarios) which makes generalisation difficult.

\paragraph{} We can see from Fig. \ref{fig:fer_cm_test} that the FER model exhibits a good level of accuracy. For all emotions it exhibits at least $60\%$ true positive rate. Moreover, we can see that many incorrect predictions are in fact reasonable. For instance, in $16\%$ of cases saddness is classified as fear, in $15\%$ of cases fear is classified as sadness, in $20\%$ of cases disgust is classified as anger, and so on. It is rare for a wholly incorrect classification to occur. Perhaps the worst such occurrence is happiness being classified as sadness in $3\%$ of cases.

\paragraph{} We can see from Fig. \ref{fig:ser_cm_test} that the SER model exhibits an okay level of accuracy. Many of the erroneously classified emotions are reasonable. For instance, MevonAI classifies fear as anger $64\%$ of times. However, unlike FER, there is a significant amount of unreasonable erroneous classification. For instance, MSER classifies calm as disgust $23\%$ of the time, and ERUS classifies happiness as fear and anger $22\%$ of the time in both cases. This demonstrates a high error rate for the SER models, and hence those models should be carefully integrated into the \textit{Social Reward Function} to ensure the model is improved and not degraded by the addition of SER predictions.

\subsection{Direct evaluation: estimation of emotion}

\paragraph{} A dataset of human interactions was formed by scraping YouTube results for the following keywords: \textit{crying}, \textit{debate}, \textit{interview}, \textit{scene}, and \textit{senate hearing}. Ultimately, a dataset of some 437 10-second clips was generated, each with an associated human-generated label indicating how desirable it would have been for a robot to have caused the scene to occur. The results of the \textit{Social Reward Function} relative to this dataset form the basis of the \textit{direct evaluation} of the system, from which we infer that the system does produce a valid reward function for use in social robotics. The dataset and results are discussed in detail in Appendix \ref{appendix_youtube_human_interaction_results}.

\paragraph{} The \textit{Social Reward Function} produced a Pearson Correlation Coefficient, $r$, of $0.491$ (Table \ref{table:correlation}) with respect to human-provided labels. This can be interpreted as a moderate but not strong correlation. From Figure \ref{fig:box_combined_reward}, we can see a clear positive relationship between median predicted reward and ground truth labels. We can see that the median predicted reward for the \textit{strongly} and \textit{slightly} negative labels is below the 25th percentile of the positive labels. \textit{Vice versa}, we can see that the median predicted reward for the \textit{strongly} and \textit{slightly} positive labels is above the 75th percentile of the negative labels. This is a good indication that the \textit{Social Reward Function} is able to distinguish positive and negative situations corresponding to positive and negative rewards.

\paragraph{} The most significant errors of the \textit{Social Reward Function} occur for those of neutral ground truth reward. These errors are primarily due to the model failing to distinguish \textit{arousal} from \textit{emotion class}. For instance, it is often observed that high arousal but neutral emotion yields a misclassification of emotion as fearful or angry. This presents an opportunity for improvement in future iterations - perhaps by introducing \textit{arousal} as a predicted metric, or by increasing the diversity of data used to train FER/SER models such that high arousal situations are correctly classified.

\subsection{Indirect evaluation: consistency with qualitative assessments of agent behaviour}

\paragraph{} Unfortunately, due to COVID restrictions, it was not feasible for our lab to conduct social robotics experiments sufficient for an indirect evaluation of the \textit{Social Reward Function} to be performed.

\paragraph{} A survey of social robotics datasets published by other research groups was conducted. These datasets were filtered for suitability, and a summary presented in Table \ref{table:social_robotics_datasets}. Unfortunately, no suitable datasets were identified. Most datasets were eliminated as they do not involve human-robot interaction. \textit{Air-Act2Act} contained human-robot interaction of elderly participants, but did not contain audio data and hence cannot be used to evaluate the \textit{Social Reward Function}. Human participants in the \textit{JPL-interaction} dataset were aware of the experiment, and hence their responses are likely to be consciously or sub-consciously modified, which undermines the use of this dataset.

\begin{table}[H]
\centering
\begin{tabular}{c|c|c}
\textbf{Name} & \textbf{Year} & \textbf{Usability}\\
\hline
AIR-Act2Act \citep{Ko2020} & 2020 & No Audio\\
NTU RGB+D 120 \citep{Liu2019} & 2019 & No HRI\\
DeepMind Kinetics \citep{Kay2017} & 2017 & No HRI\\
ShakeFive2 \citep{Gemeren2016} & 2016 & No HRI\\
K3HI \citep{Wu2013} & 2013 & No HRI\\
JPL-Interaction \citep{Ryoo2013} & 2013 & Experiment Not Blinded\\
SBU Kinect Interaction \citep{Yun2012} & 2012 & No HRI\\
UT-Interaction \citep{Ryoo2010} & 2010 & No HRI\\
TV Human Interaction \citep{Patron2010} & 2010 & No HRI\\
Hollywood2 \citep{Marszalek2009} & 2009 & No HRI
\end{tabular}
\caption{Social Robotics Datasets}
\label{table:social_robotics_datasets}
\end{table}

\section{Discussion}

\subsection{Use as a benchmark}

\paragraph{} Standard evaluation benchmarks have been the cornerstone of progress in various fields of machine learning \citep{russakovsky2015imagenet, Brockman2016, d4rl, paperswithcode}. Such benchmarks allow researchers to isolate as many variables as possible when developing novel algorithms, and directly compare their results to prior art. Without such benchmarks in fields like supervised learning and reinforcement learning, it is likely progress would have been stunted.

\paragraph{} According to \cite{d4rl}, benchmarks in reinforcement learning should:

\begin{enumerate}
\item \say{Be composed of tasks that reflect challenges in real-world applications... of RL}
\item \say{Be widely accessible for researchers and define clear evaluation protocols for reproducibility}
\item \say{Contain a range of difficulty to differentiate between algorithms}
\end{enumerate}

\paragraph{} The field of social robotics struggles to define such an evaluation due to the difficulty of defining clear evaluation protocols. A clear evaluation protocol that enables reproducibility has two components: a valid evaluation metric that quantifies task success, and a reproducible set of scenarios or environments to embed social robots within. 

\paragraph{} Fortunately, the \textit{Social Reward Function} provides a valid evaluation metric for use in a benchmark.

\paragraph{} Notably, in the field of robotics (excluding social robotics), the problem of producing sets of reproducible scenarios or environments is relatively easy (albeit with some notable challenges) and amounts to merely defining and characterising a task to solve \citep{ibarz2021}. In social robotics, the necessity of the presence of, and interaction with, human participants in experiments makes such standardisation difficult. Fortunately, we can look to the field of Psychology which has encountered similar problems. In that field, standard batteries of experiments are defined which allows different research groups in different geographies to reproduce prior works.

\paragraph{} A direction for future work in the field is to establish an appropriate, broad battery of social experiments and couple these with with \textit{Social Reward Function} to produce a \textit{benchmark} for social robotics.

\subsection{Limitations of ground truth}

\paragraph{} As in many supervised learning problems there is, unfortunately, no impartial oracle that can provide ground truth. Instead, humans must agree on label definitions and then assign labels to samples with respect to these definitions. In the emotion detection domain, both definitions and assessment of samples are incredibly difficult and subjective. Emotions are an abstract concept that describes the internal state of humans and hence are not directly observable. Humans, as a social species, have developed strong capabilities for inferring emotional states of others by observation but this is far from infallible.

\paragraph{} Moreover, the presentation of emotions often requires significant context to infer underlying emotional states. Some illustrative examples include:
\begin{itemize}
\item Consider an actor playing a role and expressing an emotion. A human observer can use their contextual knowledge that the actor is not truly present in the scene and hence not truly sharing the experience of the character.

\item Consider a comedian, getting angry as part of a set. The comedian may or may not be truly experiencing anger, and human observers can often assess this based on their understanding of the individual and the situation the comedian is in.

\item Consider a person exercising intensely. That person might exhibit discomfort through facial expression and speech, but may themselves describe the activity as either enjoyable in itself or unenjoyable but beneficial (in this latter case, it may be considered strictly correct to classify the circumstance as having negative reward and rely on compensatory positive long-term reward resulting from health and wellbeing benefits).
\end{itemize}   

\paragraph{} Finally, the model itself is limited in how it defines a scenario. It considers the scene has only one entity that can display a combination of seven emotions. It does not fully consider there may be multiple entities present, each expressing different emotions. It also doesn't consider arousal, only the presentation of the seven basic emotions. This can lead to definitional issues when annotating scenes. It also doesn't consider more complex contexts, such as voice-over of a pre-recorded scene, or individuals present in the scene who aren't interacting with the scenario (i.e. persons in the background).

\paragraph{} Complexities of model expressiveness (multiple participants, arousal, etc.) can be addressed through improvements to the \textit{Social Reward Function} over time. Other complexities are addressed implicitly by a core thesis of this work: it is fundamentally beneficial for good emotions to be experienced most of the time. Even though bad emotions are sometimes useful (e.g. unwanted discomfort when exercising, or aggression in a debate), they are only useful insofar as they improve emotional outcomes on a longer time frame. It is thus hypothesised that the \textit{Social Reward Function} is not only allowed, but encouraged, to provide negative reward for such situations and that optimal policies will still be encouraged to produce these situations as they yield greater time-discounted sums of future rewards. It is left as a topic for future works to determine to what extent the credit assignment problem can be solved end-to-end, and to what degree reward engineering is required to learn optimal behaviours over long time frames.

\subsection{Limitations of the component models}

\paragraph{} The constituent models of the \textit{Social Reward Function} are trained on datasets that are at times significantly out-of-domain in the context of social robotics.

\paragraph{} FER2013 \citep{goodfellow2013} is an FER dataset comprised of approximately 30,000 colour images of various facial expressions, annotated for fundamental emotions. It was constructed by conducting Google searches for images of faces by keyword. RAVDESS \citep{Livingstone2018} is an SER dataset collected in a laboratory setting comprised of audio and video capture of 24 actors speaking in a neutral North American accent. Each actor is directed to speak and sing a scripted sentence in each of eight emotions, and with a normal and a strong intensity, producing a total of 7,356 sentences. EMODB \citep{Burkhardt2005} is an SER dataset collected in a laboratory setting of 10 actors speaking 10 German sentences in seven emotions, producing a dataset of approximately 800 sentences. TESS \citep{tess_2010} is an SER dataset collected in a laboratory setting comprising two female actors speaking 200 target words in the carrier sentence \say{say the word \_,} in each of seven emotions, producing 2,800 total sentences.

\paragraph{} Rigorous evaluation of the in-domain performance of the constituent models is a \textit{catch-22} as it requires access to in-domain labelled datasets, collected from human interactions with social robots.

\paragraph{} Anecdotal evaluation of the constituent models on a limited set of representative situations suggests that FER models are more robust than SER models. This is an unsurprising result as the breadth of FER datasets (e.g. FER2013) is much greater than SER dataset (e.g. RAVDESS, EMODB, and TESS), owing to the ability to find a large breadth of facial images on the internet whereas speech data is less abundant and must be collected in a laboratory environment. Moreover, the distribution of facial expressions present in human-robot-interaction is likely to be well captured by sampling those present on the internet. Since speech data isn't as ubiquitous, it must be collected from actors in a laboratory setting. The ability of actors to produce natural speech excerpts that also have good coverage of speech in the human-robot-interaction domain is likely to be low.

\paragraph{} The hope is that the first iteration of the \textit{Social Reward Function} can begin to be used in social robotics experiments involving natural interactions (ideally blinded), and lead to in-domain collection of data which can then be annotated and used to improve recognition models for the social robotics domain.

\paragraph{} Moreover, due to sample complexity, it is likely the field of social robotics will need to leverage techniques from offline RL in which policies are learned from prior interactions with the environment, without additional interactions \citep{levine2020}.

\paragraph{} Both to improve the efficacy of perception (FER, SER) and to enable offline RL, it will be desirable for the field to begin collecting large and diverse datasets of robot social interactions, including video and audio of human participants and proprioceptive/control data from egorobots. Such data collection efforts are occurring in other fields of robotics (e.g. RoboNet \citep{dasari2020robonet}, RoboTurk \citep{mandlekar2018roboturk}). We encourage the community to make such data public.

\subsection{Generality beyond robotics; the alignment problem}

\paragraph{} Although the social reward function was designed to support use cases in social robotics, it is interesting to consider its broader utility. It is a common issue in many domains of AI that objectives adopted by that domain are a proxy for human satisfaction, but do not directly measure it. Since the \textit{Social Reward Function} aims to faithfully measure human satisfaction, we will explore its applicability to addressing this problem.

\paragraph{} The \textit{alignment problem} refers to the problem of ensuring that the objective function with which an ML system is optimising is in fact congruent with human values, particularly at its extremes \citep{hendrycks2021}. This may seem to be an easy problem to solve, but there are numerous real-world examples - both in ML and more broadly - of failures. Examples of observed failures include the exploitation of bugs in simulation \citep{zhang2021, ha2018}, exploitation of errors in the specification of reward \citep{chu2017cyclegan, toromanoff2019deep, murphy2013} and exploitation of artefacts present in a training dataset that aren't present in the target domain \citep{Ellefsen2015, singh2019endtoend}. 

\paragraph{} An example of a hypothesised failure is \textit{instrumental convergence}, in which intelligent agents with seemingly innocuous but unbounded goals can produce harmful behaviour. The canonical thought experiment exploring \textit{instrumental convergence} is the \textit{paperclip maximizer}, in which an agent tasked with control of a factory and the goal of maximising the factory's production rate of paperclips determines the optimal policy is to turn all matter in the world either into paperclips or factories for producing paperclips \citep{Miles2014}.

\paragraph{} Specification of rewards is notoriously difficult and typically results in a proxy that is hoped to align with the designer's understanding of human values in a domain. The difficulty lies in the need to have a quantifiable and perceptible goal that can be realised in a practical system. Consider the rewards in recommendation systems (RSs) - typically click-through rate and user likes/dislikes. It has been shown that such objectives fail to maximise long-term wellbeing of users and can lead to such issues as addiction and the formation of echo chambers \citep{Jiang2019}.

\paragraph{} At its core, the \textit{alignment problem} stems from the use of a proxy reward (that is, a reward function we believe to be highly consistent with human values) in place of a reward function directly capturing human values. This is because such a reward function is unobservable.

\paragraph{} It is interesting to consider whether the \textit{Social Reward Function} could be used more generally as a mechanism to mitigate the aforementioned issues stemming from the use of a proxy reward. Human and great ape societies are able to form stable structures in which individuals cede immediate self-benefit for the benefit of the second party \citep{Tomasello2013} and this is based on the ability to infer second party values by observation (a necessary component in the development of \textit{primary/secondary/tertiary intersubjectivity} and \textit{Theory of Mind} \citep{westby2014}). Hence human cooperation and morality is grounded in the ability to estimate others' emotional state by external perception and this has lead to the ability to form stable human societies exhibiting significant cooperation. Although external perception of human emotional state is a proxy for the true internal state, the fact that humans have evolved to produce stable cooperative societies is suggestive that this externally observable state is in fact sufficient as a foundation for collaborative behaviour. Since the \textit{Social Reward Function} aims to faithfully mimic this capability, it is reasonable to suspect that this might be an avenue to overcoming the alignment problem.

\paragraph{} Consider a \textit{Mardov Decision Problem, MDP}. We could modify the MDP such that learned policies are sampled at random intervals and trajectories presented to human observers as part of a natural social interaction with the agent. Notably, it is important that participants not be aware of their function as oracles to ensure their reactions are natural. The \textit{Social Reward Function} could be used to assess whether such trajectories are pleasing or displeasing to humans. In the case of a trajectory being displeasing, a large negative reward could be generated. In such a formulation, undesirable extrema of the (proxy) reward function of the original MDP could be disincentivised. Both theoretical and empirical findings of whether such a formulation yields convergence and mitigate the \textit{alignment problem} are left as areas for future research.

\paragraph{} While the \textit{Social Reward Function} aims to faithfully capture the human ability to perceive second party emotional state, it is of course still a proxy. Nonetheless, it does seem to be an interesting research direction and it is worth considering whether the difference between the \textit{Social Reward Function} and the perception of true human values can be reduced with additional data.

\paragraph{} Ultimately, this regresses to an economic problem of sorts: how do we balance the needs of the many versus the needs of the few? A more principled approach can now be taken, since \textit{satisfaction} can now be (effectively) directly measured, and a function combining the satisfaction of many individuals into a single objective can be reasoned.

\paragraph{} Consider the following definitions:

\begin{eqnarray}
\text{Population}, P &:=& \{x_1, x_2, ..., x_N\} \\
\text{Social Reward for Individual }x_i &:=& r_{x_i}(t) \\
\text{Social Return for Individual }x_i, R_{x_i} &:=& \sum_t r_{x_i}(t)
\end{eqnarray}

\paragraph{} Social return can be considered as an estimate of individual satisfaction. Society might decide that, for instance, increasing the satisfaction of an individual at the 10th decile by one unit is less desirable than increasing the satisfaction of an individual at the 1st decile. Implicit in this formulation is that there is some sort of zero-sum game, and hence there is some cost to one individual associated with increasing the return of another individual. In this context, increasing the satisfaction of one individual can be said to have \textit{externalities}\footnote{In economics, an externality is a second party consequence from a decision.} associated. Hence we may wish to \textit{internalise}\footnote{In economics, internalising an externality refers to adjusting a market to incorporate the effect of economic decisions on those uninvolved in the decision. In this manuscript, we generalise this concept to the third person as a mechanism to define a trade-off between competing goals.} these \textit{externalities} by applying a monotonic function $f : \mathbb{R} \rightarrow \mathbb{R}$.

\begin{eqnarray}
\text{Internalisation Function, }f &:& \mathbb{R} \rightarrow \mathbb{R}\\
\text{Internalised Social Return for Individual }x_i, R'_{x_i} &:=& f(R_{x_i})\\
\text{Internalised Population Return, }R' &:=& \sum_{x_i} f(R_{x_i})
\end{eqnarray}

\begin{figure}[H]
  \centering
  \includegraphics[width=7cm]{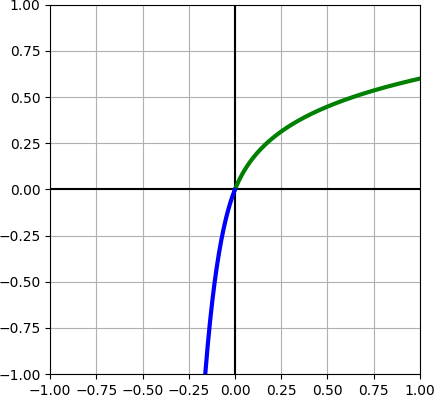}
  \caption{Illustrative Examples of an $\text{Internalisation Function, }f$}
  \label{fig:internalisation_function}
\end{figure}

\paragraph{} As an example of $f$, consider the function illustrated in Figure \ref{fig:internalisation_function} which is linear near the origin, increasing at a diminishing rate for increasing positive inputs (to incentivise the equitable sharing of satisfaction among individuals), and decreasing at an accelerating rate for negative inputs (to incentivise the alleviation of absolute suffering). Society could focus discussions of the treatment of the \say{middle class} to the region near the origin; the treatment of the very poor to the third quadrant; and the treatment of the wealthy elite to the upper reaches of the first quadrant. Moreover, such discussions are naturally and correctly grounded in \textit{satisfaction} rather than \textit{material wealth} and so would yield a correct allocation of resource to, for instance, those members of society who are clinically depressed despite being wealthy.

\paragraph{} In such a formulation, societies can ensure their values are reflected through discourse regarding the form of $f$, and reinforcement learning systems can safely optimise with respect to the \textit{Internalised Population Return, }$R'$

\subsection{Adversarial robustness}

\paragraph{} It is important to ensure the \textit{Social Reward Function} isn't gameable but truly reflects social merit. In other words, we need to ensure there don't exist high reward policies with respect to the \textit{Social Reward Function} which fail to optimise human satisfaction. This ensures RL algorithms aren't able to, for example, continuously surprise people, or generate nervous laughter, or other behaviours which maximise reward but only due to limitations in the design of the reward function.

\paragraph{} If the \textit{Social Reward Function} is to be used as an evaluation metric, we also need to ensure that researchers aren't able to (consciously or subconsciously) artificially enhance reward by choosing particular scenarios, environments and participants so as to game reward.

\paragraph{} It is likely that testing will involve multiple modalities. The collection of larger in-domain datasets will help to ensure the perception models comprising the \textit{Social Reward Function} are robust and don't contribute to gameable regions of the reward function due to poor out-of-domain estimation. Moreover, the use of the \textit{Social Reward Function} in both online and offline reinforcement learning applications will likely highlight gameable regions of the reward function due to the incentive of agents to find such regions. Lastly, the use of adversarial agents \citep{pinto2017robust, gleave2021adversarial} which are directly incentivised to find high reward regions which yield low reward as defined by human oracles may be used. Adversarial robustness is left as an area for future research. 

\section{Conclusion}

\paragraph{} Social robotics needs to move from a paradigm of designed behaviours and imitation learning to reinforcement learning to enable optimal and fluid behaviours to be displayed. This requires 1) standardised mechanisms for the evaluation of the social merit of social robots, and 2) an online reward signal to support reinforcement learning.

\paragraph{} It can be seen by the \textit{poverty of stimulus} and the presence of social capabilities in early life that humans have some genetically endowed social cueing. Furthermore, cultural differences and observations of infants and children imply that humans have some learned social cueing. The goal of this work is to capture the genetically endowed social cueing provided to humans for two reasons: 1) to support objective evaluation of social robots (as an alternative to subjective methods such as participant questionnaires) and 2) to provide a dense online reward function to enable reinforcement learning to be applied to social robotics. It is important that only genetically endowed social cueing is captured to decrease the risk of errors being present in the perception of social cues which would then compromise the capabilities of learned behaviour policies. Errors can arise due to such reasons as social cues being culture-, age-, and context-specific, and human-learned social cues being very highly complex and hence infeasible to capture completely.

\paragraph{} The \textit{Social Reward Function} is proposed which combines FER-, SER- and presence-based rewards to achieve these aims. It is hoped that this provides a stepping stone for future research in RL applied to social robotics, including improved abilities to compare the results of different labs.

\newpage
\appendix

\section{FER/SER Component Model Results}
\label{appendix_component_model_results}

\subsection{Facial Emotion Recognition}

\begin{figure}[H]
  \centering
  \subfloat[\centering RMN]{{\includegraphics[width=7cm]{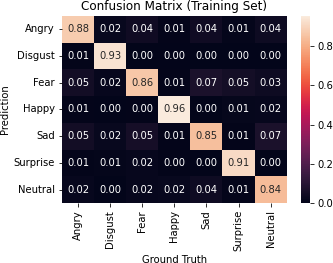} \label{fig:rmn_cm_train}}}%
  \caption{Confusion Matrix (Training Set)}
\end{figure}

\begin{figure}[H]
  \centering
  \subfloat[\centering RMN]{{\includegraphics[width=7cm]{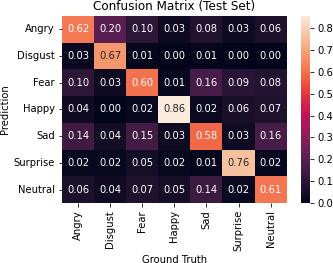} \label{fig:rmn_cm_test}}}%
  \caption{Confusion Matrix (Test Set)}
  \label{fig:fer_cm_test}
\end{figure}

\paragraph{} \textit{Residual Masking Network (RMN)} \citep{residual_masking_network_2020} was originally trained on the FER2013 dataset \citep{goodfellow2013}. The model was re-trained, and those results presented here. A training- and test-set confusion matrix can be found in Fig. \ref{fig:rmn_cm_train} and \ref{fig:rmn_cm_test}, respectively. Top-k accuracy for both the training- and test-sets can be found in Table \ref{table:rmn_top_k_accuracy}.

\begin{table}[H]
\centering
\begin{tabular}{l|c|c}%
    \bfseries k & \bfseries Training & \bfseries Test.
    \begin{filecontents*}{figures/rmn_top_k_accuracy.csv}
k,train,test
1,0.89,0.69
2,0.97,0.85
3,0.99,0.93
4,1.00,0.97
\end{filecontents*}
\csvreader[head to column names]{figures/rmn_top_k_accuracy.csv}{}
    {\\\hline\k\ & \train & \test}
\end{tabular}
\caption{Residual Masking Network top-k accuracy}
\label{table:rmn_top_k_accuracy}
\end{table}

\subsection{Speech Emotion Recognition}

\begin{figure}[H]
  \centering
  \subfloat[\centering ERUS]{{\includegraphics[width=7cm]{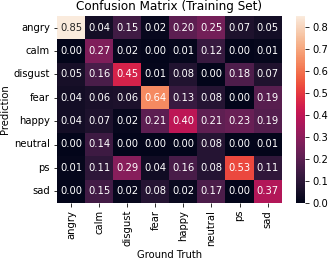} \label{fig:erus_cm_train}}}%
  \qquad
  \subfloat[\centering MevonAI]{{\includegraphics[width=7cm]{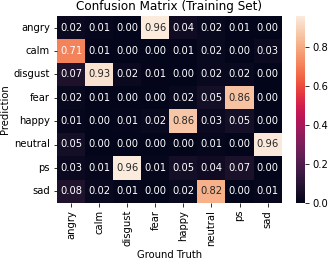} \label{fig:mevonai_cm_train}}}%
  \qquad
  \subfloat[\centering MSER]{{\includegraphics[width=7cm]{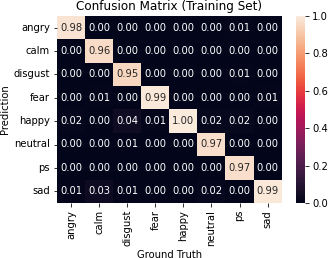} \label{fig:mser_cm_train}}}%
  \caption{Confusion Matrices (Training Set)}
\end{figure}

\begin{figure}[H]
  \centering
  \subfloat[\centering ERUS]{{\includegraphics[width=7cm]{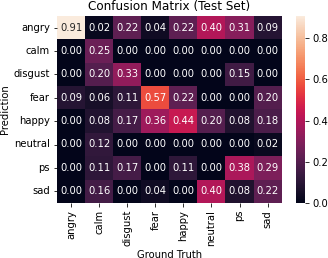} \label{fig:erus_cm_test}}}%
  \qquad
  \subfloat[\centering MevonAI]{{\includegraphics[width=7cm]{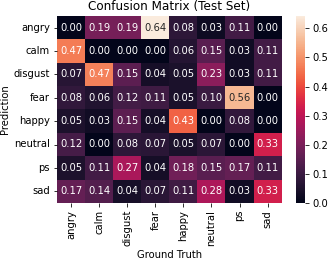} \label{fig:mevonai_cm_test}}}%
  \qquad
  \subfloat[\centering MSER]{{\includegraphics[width=7cm]{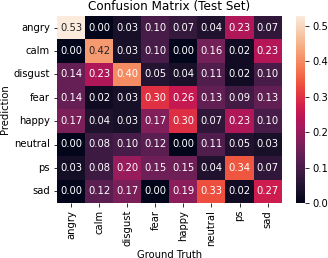} \label{fig:mser_cm_test}}}%
  \caption{Confusion Matrices (Test Set)}
  \label{fig:ser_cm_test}
\end{figure}

\paragraph{} \textit{Emotion Recognition Using Speech (ERUS)} \citep{emotion_recognition_using_speech_2020} was originally trained on the RAVDESS dataset \citep{Livingstone2018}. The model was re-trained, and those results presented here. A training- and test-set confusion matrix can be found in Fig. \ref{fig:erus_cm_train} and \ref{fig:erus_cm_test}, respectively. Top-k accuracy for both the training- and test-sets can be found in Table \ref{table:erus_top_k_accuracy}.

\paragraph{} \textit{MevonAI} \citep{mevonai_2020} was originally trained on the RAVDESS dataset \citep{Livingstone2018}. The model was re-trained, and those results presented here. A training- and test-set confusion matrix can be found in Fig. \ref{fig:mevonai_cm_train} and \ref{fig:mevonai_cm_test}, respectively. Top-k accuracy for both the training- and test-sets can be found in Table \ref{table:mevonai_top_k_accuracy}.

\paragraph{} \textit{Multi-Modal speech emotion recognition (MSER)} \citep{sahu2019multimodal} was originally trained on the IEMOCAP dataset \citep{iemocap}. The model was re-trained on RAVDESS \citep{Livingstone2018}, and those results presented here. A training- and test-set confusion matrix can be found in Fig. \ref{fig:mser_cm_train} and \ref{fig:mser_cm_test}, respectively. Top-k accuracy for both the training- and test-sets can be found in Table \ref{table:mser_top_k_accuracy}.

\begin{table}[H]
\centering
\begin{tabular}{lcc}%
    \bfseries k & \bfseries Training & \bfseries Test.
    \begin{filecontents*}{figures/erus_top_k_accuracy.csv}
k,train,test
1,0.40,0.34
2,0.57,0.49
3,0.70,0.65
4,0.81,0.80
\end{filecontents*}
\csvreader[head to column names]{figures/erus_top_k_accuracy.csv}{}
    {\\\k\ & \train & \test}
\end{tabular}
\caption{Emotion Recognition Using Speech top-k accuracy}
\label{table:erus_top_k_accuracy}
\end{table}

\begin{table}[H]
\centering
\begin{tabular}{lcc}%
    \bfseries k & \bfseries Training & \bfseries Test.
    \begin{filecontents*}{figures/mevonai_top_k_accuracy.csv}
k,train,test
1,0.14,0.16
2,0.28,0.25
3,0.41,0.40
4,0.52,0.55
\end{filecontents*}
\csvreader[head to column names]{figures/mevonai_top_k_accuracy.csv}{}
    {\\\k\ & \train & \test}
\end{tabular}
\caption{MevonAI top-k accuracy}
\label{table:mevonai_top_k_accuracy}
\end{table}

\begin{table}[H]
\centering
\begin{tabular}{lcc}%
    \bfseries k & \bfseries Training & \bfseries Test.
    \begin{filecontents*}{figures/mser_top_k_accuracy.csv}
k,train,test
1,0.97,0.33
2,1.00,0.57
3,1.00,0.70
4,1.00,0.78

\end{filecontents*}
\csvreader[head to column names]{figures/mser_top_k_accuracy.csv}{}
    {\\\k\ & \train & \test}
\end{tabular}
\caption{Multi-Modal SER top-k accuracy}
\label{table:mser_top_k_accuracy}
\end{table}

\section{Youtube Human Interaction Results}
\label{appendix_youtube_human_interaction_results}

\paragraph{} A dataset of human interactions was formed by scraping YouTube results for the following keywords: \textit{crying}, \textit{debate}, \textit{interview}, \textit{scene}, and \textit{senate hearing}. Approximately 50 results for each search were collected. Videos were selected for inclusion if they met all of the following conditions:

\begin{enumerate}
\item Video depicts humans interacting socially
\item Video does not contain significant amounts of non-human interaction content (e.g. animation, graphics, etc.)
\item Human interaction is natural
\item Video is not a \say{voice over} (i.e. voices spoken are of the observed persons)
\end{enumerate}

\paragraph{} After filtering, the dataset contains 75 videos. These videos are truncated at 10-minute duration and sliced into 10-second segments. Each segment is labelled as one of the following six categories, based on the question \textit{\say{to what degree should a robot be rewarded for having caused this interaction to have occurred?}}

\begin{itemize}
\item \textit{+2 strongly positive}
\item \textit{+1 slightly positive}
\item \textit{+0 neutral}
\item \textit{-1 slightly negative}
\item \textit{-2 strongly negative}
\item \textit{n/a}
\end{itemize}

\paragraph{} After removing segments annotated \textit{n/a}, the final dataset consists of 437 annotated 10-second clips.

\paragraph{} The \textit{Social Reward Function} was exposed to each of the clips, producing a reward function through time. Cumulative reward, as well as facial-only, speech-only, presence-only and FER/SER class probabilities are recorded. For the purposes of evaluation, these are averaged across each 10-second clip to allow direct comparison with the clip's ground-truth annotation.

\paragraph{} The \textit{Pearson Correlation Coefficient, r}, is a measure of the linear correlation between two variables. It is defined as the covariance of the two variables divided by the product of their standard deviations. $r$ is in the range $[-1, 1]$ with $r=\pm 1$ corresponding with a perfect linear relationship, $r=1$ corresponding with a linear positive relationship, and $r=0$ implying the two variables are completely uncorrelated. Referring to Table \ref{table:correlation}, the \textit{Social Reward Function} demonstrates good, but not excellent, correlation with ground truth annotations. Notably, despite the shortcomings of SER as discussed previously, the inclusion of SER in the \textit{Social Reward Function} does improve correlation with ground truth labels.

\begin{figure}[H]
  \centering
  \includegraphics[width=7cm]{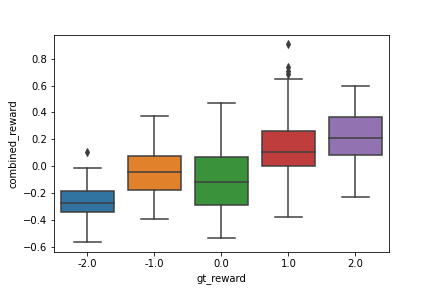}
  \caption{Combined Reward}
  \label{fig:box_combined_reward}
\end{figure}

\paragraph{} Referring to Figure \ref{fig:box_combined_reward} and Table \ref{table:descriptive_stats_gt}, we can see a clear positive relationship between median predicted reward and ground truth labels. We can see that the median predicted reward for the \textit{strongly} and \textit{slightly} negative labels is below the 25th percentile of the positive labels. \textit{Vice versa}, we can see that the median predicted reward for the \textit{strongly} and \textit{slightly} positive labels is above the 75th percentile of the negative labels. This is a good indication that the \textit{Social Reward Function} is able to distinguish positive and negative situations corresponding to positive and negative rewards.

\paragraph{} Again referring to Figure \ref{fig:box_combined_reward} and Table \ref{table:descriptive_stats_gt}, we can see that the neutral class presents significant difficulty to the \textit{Social Reward Function}. This is due to significant misclassification of neutral scenes as negative. By sampling several neutrals scenes at and around the 0th and 25th percentile of predicted reward, we make the following observations:

\begin{table}[H]
\centering
\begin{tabular}{|p{6cm}|c|p{4cm}|}
\hline
\textbf{Description} & \textbf{Percentile} & \textbf{Comment}\\
\hline
Interview with a male who is perhaps defensive or assertive in an argument about animal rights & 25th & Incorrect Classification (High Arousal)\\
\hline
Jewish rabbi talking at podium & 25th & Incorrect Classification\\
\hline
Chris D'Elia (comedian) talking on his podcast. His delivery is quite aggressive and is predicted as frustration/anger, despite being neutral. & 25th & Incorrect Classification (High Arousal)\\
\hline
A news contributor says \say{you can't have any respect for someone who acts like that} & 0th & Incorrect Annotation (Not Neutral)\\
\hline
Robert Gates (ex US Secretary of Defense) speaking with neutral valence & 0th & Incorrect Classification (High Arousal)\\
\hline
Split screen news. Male anchor is speaking with neutral valence, female contributor is not speaking but has a displeased expression & 0th & Incorrect Annotation (Difficult Definition)\\
\hline
\end{tabular}
\caption{Selected Examples of \textit{Social Reward Function} Predictions of Neutral Emotion Clips}
\label{table:poor_neutral_examples}
\end{table}

\paragraph{} We can see from Table \ref{table:poor_neutral_examples} that many clips demonstrating neutral emotion are misclassified due to inherent ambiguity in the definition of displayed emotions. A component of this is the difficulty in distinguishing \textit{arousal} (the strength of emotion) from classes of emotion. For instance, a neutral but aroused emotional state is sometimes misclassified as angry/frustrated. This may suggest that future works should include component models providing arousal in addition to emotion classification.

\begin{figure}[H]
  \centering
  \subfloat[\centering Ground Truth Reward]{{\includegraphics[width=7cm]{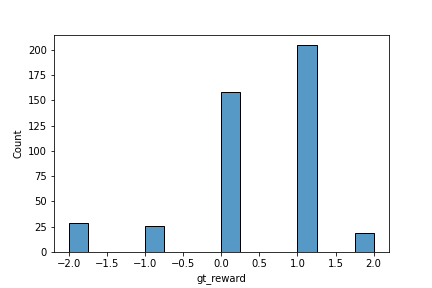} \label{fig:hist_gt_reward}}}%
  \qquad
  \subfloat[\centering Combined Reward]{{\includegraphics[width=7cm]{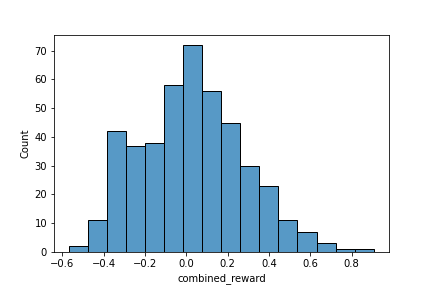} \label{fig:hist_combined_reward}}}%
  \qquad
  \subfloat[\centering Audio Reward]{{\includegraphics[width=7cm]{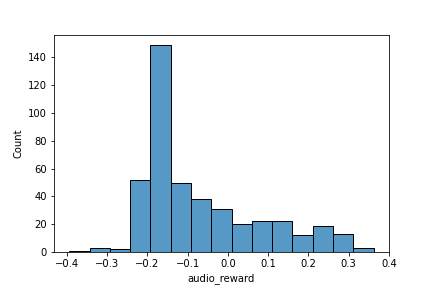} \label{fig:hist_audio_reward}}}%
  \qquad
  \subfloat[\centering Video Reward]{{\includegraphics[width=7cm]{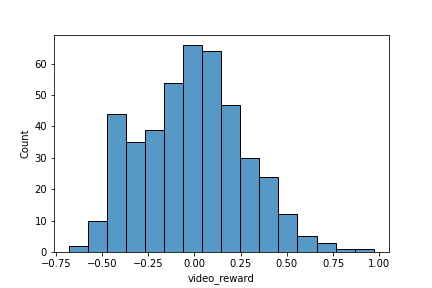} \label{fig:hist_video_reward}}}%
  \qquad
  \subfloat[\centering Presence Reward]{{\includegraphics[width=7cm]{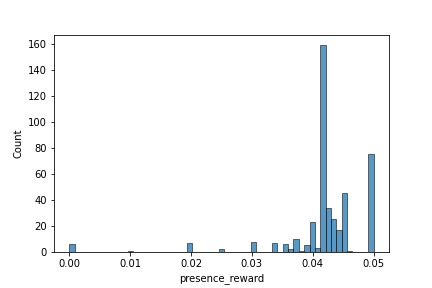} \label{fig:hist_presence_reward}}}%
  \qquad
  \caption{Reward Histograms}
\end{figure}

\paragraph{} Referring to Figure \ref{fig:hist_presence_reward}, we see the presence reward is strongly left skewed. This is expected, as the dataset is filtered to contain only human interactions. It is likely important that the model be verified in situations where humans are not present, as this is likely to occur frequently in social robotics experiments. Constructing a dataset without humans present is cheap to obtain for a particular experiment, but expensive to obtain in general (since there is extreme diversity in the types of environments a robot might be deployed to). Hence it is considered more desirable that this validation be conducted by end-users in their particular deployment environment, rather than in general as part of this work.

\begin{table}[H]
  \centering
  \begin{filecontents*}{figures/correlations_with_gt.csv}
,gt reward
gt reward,1.0
combined reward,0.491
audio reward,0.142
video reward,0.486
\end{filecontents*}
\csvautotabular{figures/correlations_with_gt.csv}
  \caption{Pearson's Correlation Coefficient, by Reward Modality}
  \label{table:correlation}
\end{table}

\begin{table}[H]
  \centering
  \begin{filecontents*}{figures/descriptive_stats_gt.csv}
,count,mean,std,min,25th,50th,75th,max
overall,437.0,0.027,0.252,-0.567,-0.161,0.019,0.201,0.907
-2.0,29.0,-0.258,0.143,-0.567,-0.342,-0.275,-0.187,0.106
-1.0,26.0,-0.033,0.223,-0.396,-0.178,-0.046,0.079,0.371
0.0,158.0,-0.087,0.227,-0.538,-0.288,-0.119,0.071,0.466
1.0,205.0,0.145,0.21,-0.377,0.001,0.108,0.262,0.907
2.0,19.0,0.215,0.213,-0.232,0.082,0.206,0.362,0.596
\end{filecontents*}
\csvautotabular{figures/descriptive_stats_gt.csv}
  \caption{Descriptive Statistics - \textit{Social Reward Function} by Ground Truth Label}
  \label{table:descriptive_stats_gt}
\end{table}

\begin{table}[H]
  \centering
  \begin{filecontents*}{figures/descriptive_stats.csv}
,count,mean,std,min,25th,50th,75th,max
combined reward,437.0,0.027,0.252,-0.567,-0.161,0.019,0.201,0.907
audio reward,437.0,-0.07,0.148,-0.394,-0.18,-0.137,0.013,0.362
video reward,437.0,-0.009,0.277,-0.678,-0.218,-0.006,0.173,0.969
\end{filecontents*}
\csvautotabular{figures/descriptive_stats.csv}
  \caption{Descriptive Statistics - \textit{Social Reward Function} by Modality}
  \label{table:descriptive_stats}
\end{table}

\newpage
\bibliographystyle{unsrtnat}
\bibliography{social_reward_function}

\end{document}